\documentclass[11pt,a4paper]{article}
\usepackage{times,latexsym}
\usepackage{url}
\usepackage[T1]{fontenc}

\usepackage[acceptedWithA]{tacl2021v1}
\usepackage{amsmath}
\usepackage{amssymb}
\usepackage{listings}
\usepackage{lipsum}
\usepackage{wrapfig} % for wrapping the image
\usepackage{xcolor} % for setting colors
\usepackage{float}
\usepackage{pgfplots}
\usepackage{forest}
\usepackage{tikz}
\usepackage{tikz-dependency}
\usepackage{xspace}
\usepackage{multirow}
\usepackage{todonotes}
\usepackage{fancyvrb}
\usepackage{overpic}
\usepackage{soul}

\pgfplotsset{compat=newest}
\pgfdeclarelayer{background}
\pgfsetlayers{background,main}
\usetikzlibrary{trees, arrows, positioning, shapes.geometric}

\lstdefinestyle{python}{
    language=Python,
    basicstyle=\ttfamily\tiny,
    numberstyle=\tiny\color{gray},
    numbers=left,
    stepnumber=1,
    firstnumber=1,
    numberfirstline=true,
    breaklines=true, 
    postbreak=\mbox{\textcolor{red}{$\hookrightarrow$}\space}
}

\lstdefinestyle{python-medium}{
    language=Python,
    basicstyle=\fontsize{5pt}{6pt}\selectfont\ttfamily,
    numbers=none,
    breaklines=true, 
    postbreak=\mbox{\textcolor{red}{$\hookrightarrow$}\space}
}

\lstdefinestyle{python-small}{
    language=Python,
    basicstyle=\fontsize{4pt}{5pt}\selectfont\ttfamily,
    numbers=none,
    breaklines=true, 
    postbreak=\mbox{\textcolor{red}{$\hookrightarrow$}\space}
}

\lstdefinestyle{txt}{
    basicstyle=\ttfamily\tiny, % Small typewriter font
    breaklines=true,            % Enable line breaking
    postbreak=\mbox{\textcolor{red}{$\hookrightarrow$}\space} % Optional: mark where lines break
}

\definecolor{deepgray}{HTML}{505050}
\definecolor{lightgray}{HTML}{A9A9A9}
\definecolor{pastelyello}{HTML}{FFD700}

\newcommand{\workTitle}{NoviCode: Generating Programs\\ from Natural Language Utterances by Novices}

\newcommand{\workNameShort}{NoviCode\xspace}

\title{\workTitle}

\author{
  Asaf Achi Mordechai \quad
  Yoav Goldberg \quad
  Reut Tsarfaty
  \\
  \\
  Computer Science Department
  \\
  Bar-Ilan University
  \\
  Ramat-Gan, Israel
  \\
  \texttt{\{asaf.achimordechai, yoav.goldberg, reut.tsarfaty\}@gmail.com}
}

\date{}

\begin{document}
\maketitle
\begin{abstract}
%% motivation -- why should we care?
Current Text-to-Code models demonstrate impressive capabilities in generating executable code from natural language snippets. However, current studies focus on technical instructions and programmer-oriented language, and it is an open question whether these models can effectively translate natural language descriptions given by non-technical users and express complex goals, to an executable program that contains an intricate flow --- composed of API access and control structures as loops, conditions, and sequences. %\yg{what is etc? doesn't look good} 
%In this work, our contribution
To unlock the challenge of generating a complete program from a plain non-technical description 
%---  without the specification of required technical concepts as would be given by a programmer --- 
we present \workNameShort, a novel %\workName 
NL Programming task, which takes as input an API and a natural language description by a novice non-programmer, and provides an executable program as output. 
To assess the efficacy of models on this task, we provide a novel benchmark accompanied by test suites wherein the generated program code is assessed not according to their form, but according to their functional execution.
Our experiments show that, first, \workNameShort is indeed a challenging task in the code synthesis domain, and 
that generating complex code from non-technical instructions %\yg{mention the novice, non-technical %instructions?} 
%and profoundly evaluating it
goes beyond the current Text-to-Code paradigm. 
Second, we show that a novel approach wherein we align the NL utterances with the compositional hierarchical structure of the code, %\yg{is this still true?}\asaf{yes, only regarding the code intermediate form} 
greatly enhances the performance of LLMs on this task, compared with the end-to-end Text-to-Code counterparts.

\end{abstract}

\section{Introduction}
\label{section:introduction}
%\yg{NOTE: some paragaph titles end with a period, and others do not. be consistent, either all with, or all without. similarly for capitalization choice in section titles.}
%% goal
%In the era of generative AI, code generation from text has become a popular and useful domain of investigation within NLP.\yg{cite! also i am not sure we want to say "within NLP", why do we need this qualification?}
The current Text-to-Code paradigm focuses on generating {\em code-lines} from technical descriptions produced by trained programmers.
In this work, we move from this Text-to-Code paradigm towards generating {\em programs} with intricate structures from intuitive, \emph{day-to-day language descriptions} produced by \emph{non-technical} individuals. 
This novel reconstruction of the task is challenging on two levels: (a) the generated programs have non-trivial control-flow structures --- with API calls to a novel API that the model was not necessarily trained on --- 
rather than generic one-liners; and 
(b) we are interested in instructions provided by laypeople in everyday natural language, without using technical concepts or jargon.  We term this task {\em Natural Language Programming} (NLProg).

\begin{figure*}
    \centering
    \begin{minipage}[t]{.45\textwidth}
    \centering
    \textbf{\quad \quad \quad \quad Input}
    \vspace{0.2cm}
    \newline
    \footnotesize
    \begin{tikzpicture}
        % API icon
        \node (person) at (0,2) {\includegraphics[width=0.5cm]{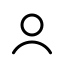}};
        \node[left=0.0cm of person] {\textbf{\quad \quad \quad User NL command}};
    \end{tikzpicture}
    \newline
    \footnotesize
    \textit{\scriptsize "Check that I received confirmation emails from all advisors in the committee or cancel my meeting with them"}
    \newline
    \begin{tikzpicture}
        \draw[dashed, line width=0.25pt, black] (-4,0) -- (4,0);
    \end{tikzpicture}
    \footnotesize
    \begin{tikzpicture}
        % API icon
        \node (api) at (0,2) {\includegraphics[width=0.5cm]{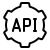}};
        \node[left=0.0cm of api] {\textbf{API Specifications}};
    \end{tikzpicture}
    \begin{lstlisting}[style=python]
class Messages(Action):
    def find_messages(
        cls,
        date_time: Optional[DateTime] = None,
        sender: Optional[Contact] = None,
        recipient: Optional[Contact] = None,
        content: Optional[Content] = None
    ) -> List[MessageEntity]:
        pass

class Calendar(Action):
    def delete_events(
        cls,
        date_time: Optional[DateTime] = None,
        event_name: Optional[EventName] = None,
        participants: Optional[List[Contact]] = None
    ) -> List[EventEntity]:
        pass

class Resolvable(Generic[T]):
    def resolve_from_text(T, text: str) -> T:
        pass

    def resolve_many_from_text(T, text: str) -> List[T]:
        pass
    \end{lstlisting}
    \end{minipage}\hfill
    \begin{minipage}[t]{.45\textwidth}
    \centering
    \textbf{\quad \quad \quad Output}
    \vspace{0.2cm}
    \newline
    \footnotesize
    \begin{tikzpicture}
        % API icon
        \node (program) at (0,2) {\includegraphics[width=0.5cm]{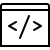}};
        \node[left=0.0cm of api] {\textbf{Program}};
    \end{tikzpicture}
    % \vspace{1cm}
    \begin{lstlisting}[style=python]
content = Content.resolve_from_text('confirmation')
senders = Contact.resolve_many_from_text('all advisors in the committee')
messages = []
for sender in senders:
    messages.append(
        Messages.find_messages(sender=sender, content=content)
    )
test_messages = all(messages)
    
if not test_messages:
    event_name = EventName.resolve_from_text('my meetings with them')
    participants = senders
    Calendar.delete_events(
        event_name=event_name,
        participants=participants
    )
    \end{lstlisting}
    \end{minipage}
    \caption{The NLProg Task. Left: The input, consisting of an utterance and an API specification for implementing the target code. Right: The output, reflecting the procedural instructions in the utterance, is implemented with the correct usage of the API specifications.}
    \label{fig:intro-example}
\end{figure*}

%% introduce code generation for novice NL desc
NLProg differs from standard code generation models,  tasked to convert natural language instructions or descriptions into executable code, in several ways.
In the Text-to-Code paradigm, models require the user to use technical terms, like flow structures (e.g., conditions or loops), data types, variables, and programming concepts (e.g., sorting algorithms, object-oriented concepts, recursions, etc.). 
In contrast, in NLProg %we aim there is an evident gap to overcome\yg{i have no idea what the previous sentence says} --- 
we aim to convert plain, everyday language descriptions, devoid of any technical jargon, into functional code using independent and unseen API specifications.

%% exemplify code generation for novice NL desc 
NLProg is particularly challenging when the descriptions implicitly allude to code constructs as control flow elements such as sequences, loops, or conditional statements, {\em without} explicitly mentioning them.
For example, given a standard API of email and calendar applications and a user request ``\textit{Check that I received confirmation emails from all advisors in the committee or cancel my meeting with them}'', we expect the model to interpret the conjunction ``\textit{or}'' as indicating a condition, and to recognize ``\textit{all advisors in the committee}'' as an iteration over a particular set. The model should then generate a proper executable program based on the API (Figure~\ref{fig:intro-example}). %Yet, code synthesis models consistently fail\rt{where do we see this?} even on toy examples such as this one.

%% what did previous ppl do in this domain 
Recent work on natural language synthesis to code (\citealp{DBLP:journals/corr/abs-2107-03374}, \citealp{wang2023codet5+}, \citealp{nijkamp2023codegen2}) have shown remarkable success in code generation tasks using LLMs.
Trained on a large fraction of GitHub repositories (\citealp{DBLP:journals/corr/abs-1808-09588}, \citealp{DBLP:journals/corr/abs-1909-09436}, \citealp{lu2021codexglue}, \citealp{wang2021codet5}, \citealp{wang2023codet5+}) or StackOverflow question-answer pairs \citep{DBLP:journals/corr/abs-1805-08949}, these code-oriented LLMs  learn rich contextual representations that can be transferred to various code-related downstream tasks.
%% challenges in previous works

However, these works share several limitations. They predominantly focus on technically precise descriptions {\em given by  programmers}, using jargon-laden, code-centric discourse geared towards seasoned programmers. Other datasets in intent recognition and slot fillings (\citealp{DBLP:journals/corr/abs-1810-07942}, \citealp{DBLP:journals/corr/abs-2010-03546}) focus on plain language description from non-programmers, or novices, but these are translated into \emph{simple code} that exhibits no complexity in its flow execution and is inherently derived from the NL descriptions in its datasets.

%% task + benchmark
To bridge this gap, %\yg{What challenges? maybe say "this gap" instead?} 
we present \workNameShort, a novel task that aims to translate NL descriptions from novices into complex executable programs. 
%
%To this end, 
We curated a dataset where novice-centric NL instructions are paired with complex code programs. These code programs adhere to API specifications to facilitate the functional execution of the produced code.
Having defined this task and created the benchmark, two additional challenges present themselves, namely: (i) How can we evaluate the efficacy of models on such complex code synthesis tasks? and (ii) What would be good models for addressing the non-trivial challenge of translating high-level language to low-level intricate code?

%% eval/quality assessment
To address the evaluation challenge, we deliver, along with the task data, test suites that simulate diverse scenarios of execution flows to gauge the functional correctness of the generated code. 
Our evaluation methodology is designed to assess the functional correctness of the control flow elements within the generated complex programs, rather than merely looking at the {\em form} of the code or assessing an {\em end state} only.

%% modeling
To address %\yg{maybe move this paragraph to where I mark with "*"?}
the modeling challenge, we hypothesize that learning the alignment of the natural language queries with codes'  hierarchical structures, rather than learning a simple end-to-end text-to-code model, can improve performance, particularly with  programs as complex as we target here. 

%% empiric study
Our experiments with standard LLMs 
assessed by our functional  evaluation suites demonstrate that the task is indeed challenging. Existing models struggle to produce the expected executable code with complex constructs. %\yg{*} 
We further empirically confirm our hypothesis that learning the alignment of natural language spans to compositional hierarchical code structures   is better than the de-facto standard  end-to-end modeling.

%% summary
The contribution of this paper is thus multifold. 
First, we define a novel code-synthesis task based on plain non-technical natural language.
Second, we deliver a dataset of NL instructions obtained from non-programmers, paired with an expert-crafted evaluation suite that focuses on the function, rather than form  of the generated code.  Finally, we provide a novel modeling approach that aligns  NL spans with the explicit compositional structures of the code. We further show that this approach outperforms the standard end-to-end baseline models on this task, even with the most capable contemporary generative LLMs.

\section{The Challenge: Technical Code from Non-Technical Users}
\label{section:challenges}

\subsection{Terminology}
\label{section:introduction}

In this work we are interested in synthesizing complex programs from descriptions by novice users. %\ygrem{, we term this natural language programming}. 
Before we begin, we define the relevant terms:
%: Natural Language Programming, Simple vs Complex programs, and Novices vs Experts.
%....\rt{We first define the terminology, and in 2.2. we contrast different combinati}
\paragraph{Natural Language Programming.}
We equate programming in natural language with the ability to translate a desired scenario using plain, everyday language to a working program \citep{harel2008can}.
%\yg{I don't understand the part about the API. Does the user know about the API? if not, why is it part of the NL Prog?}
%Often, NL programming involves an API to be accessed, and user requests to process the outcome.

\paragraph{Novices vs.\ Programmers.}
When crafting NL descriptions of programs, we distinguish  two distinct types of personas, defined by their programming skills:
%\yg{switch the order of programers and novices?}
\textbf{Programmers}, who often use technical terminology --- such as functions, loops, iterations, conditions, exceptions, methods, and  concepts that are common when expressing computations --- which is not typically familiar to individuals without formal training in the field.
\textbf{Novices}, who are devoid of programming skills, utilizing intuitive, everyday language to describe program functionality without using technical terms or referring to specific programming paradigms.
%\textbf{Programmers} leverage technical jargon and nuanced programming concepts, such as algorithms (e.g., \textit{binary search}), methodologies (e.g., \textit{authentication}), and concepts (e.g., \textit{client-server}), often referencing specific software ontology terms, like \textit{DataFrame}, \textit{port}, and \textit{file system}.
%\yg{I added the following sentence} \asaf{I used it in the previous sentence instead of the previous content we had on programmers} %Programmers also often talk in terms of programming -- functions, concepts such as loops, iterations, conditions, exceptions, methods, and similar concepts that are common when expressing computations, yet are not natural for untrained lay people.

\paragraph{Simple vs.\ Complex programs.} In this work, we define a {\bf simple} program as a simple statement that abstains from long sequences of actions or the incorporation of logical control-flow elements. 
A program is deemed {\bf complex} if it executes a sequence of statements or has one or more control flow structures, such as loops or conditions.
%Models for translating text to code are primarily designed with a programmer-centric perspective in mind. Yet, novice descriptions of programs present unique challenges: they use common language, lack technical terms, and implicitly depict elements in the code. 
%In addition, evaluating generated code is hard, and evaluating generated complex code from novice utterances is harder. Complex code requires more comprehensive and granular evaluations. Functional correctness evaluation metrics are task-specific and are not suitable for programs generated from novice utterances.\reut{in case of lack of space, consider removing this section intro and lead directly to subsec}

\subsection{Why is Programming in Natural Language  Challenging?}

%\ygrem{Having defined the unique traits of our task -- we are now set to identify key challenges in generating complex programs from novice NL utterances, and how they differ from the text-to-code task.}

% high-level challenge
Programmers and novices describe code differently   --- with programmers describing explicitly elements in the target code, like functions, arguments, and variable names. Programmers express ideas using technical jargon and programming concepts used in the code.
%\reut{The overall challenge section with both subsec should not exceed 1 page IMO}

Let's consider for example the following articulation of a {\em simple} program by a programmer: "\textit{Start a server listening on port 8080}". This example shows a close coupling of the description and the program code. This description explicitly specifies the arguments (\textit{port}) and their respective values (\textit{8080}) in the \textit{start server} API. The ontological terms (\textit{server} and \textit{port}) in the description also exemplify the technical language used by programmers. Last, the description uses code concepts like \textit{Start a server} and \textit{listen on port}.

Moreover, when describing a {\em complex} program in natural language, programmers typically use a structured language to describe the logical control flows within the desired program. In their description, specific keywords (e.g. \textit{if-else} and \textit{for} or \textit{while}) are used to describe conditionals or loops. Loops are also described using explicit quantifiers. Control flow structures, like sequences of operations, are often brought in the plain order of execution. 
For example, "\textit{Find the average of every three columns in Pandas DataFrame and then return the max value}". This description exhibits the logical flow in the order it should appear in the code -- it indicates a sequential operation, which begins in a loop and then commences with a simple call for a function (i.e. \textit{max}). The loop is described using a quantified term (i.e. \textit{every three}).

On the other end, Novices craft NL utterances unaligned with
the program code they describe. They have no prior knowledge of the target program code or programming language, the API being used, or programming concepts in general --- making their descriptions simpler using everyday language and diverging from the underlying code elements. 
When describing complex programs, novices use semantics that implicitly hints at loops or conditions. Furthermore, novices' discourse in the utterances is not constrained to the order of logical control flow structures in the code but to their more intuitive capture in their minds.

To illustrate, examine the following novice description of a complex program: "\textit{Find me adjacent seats for Shakespeare in the Park, on the first weekend day the weather will be nice}". In this example, and depending on the specific API, the expected resulting program might need to contain a loop (i.e., iterating over \textit{the weekend} days) that is contingent on the result of a condition (i.e., \textit{the weather will be nice}). One should notice that the order of the predicates in the description differs from the order of the statement executed in the code. Furthermore, no explicit keywords were used to indicate the condition or the loop. Last, the NL terms do not necessarily align with the actual arguments within the code (i.e. "\textit{weather will be nice}" will be inferred to a range of temperature and other weather features in the code). This creates an additional challenge, that of translating of language used in novices' utterances into the technical intricacies of the target code. 
%In contrast, the programmers' utterances are closer to the resultant code thus making it simpler for the model to generate the expected resultant code.

\subsection{Why is the Evaluation of NL Programming Challenging?}

Code generation models have traditionally been assessed by string-match-based metrics, such as BLEU \cite{papineni-etal-2002-bleu} and CodeBLEU \cite{DBLP:journals/corr/abs-2009-10297}. However, these methods fall short because they misjudge functionally equivalent solutions to the reference solution in the vast space of programs.
As a result, recent work in code synthesis turned to evaluating models through the functional correctness of the generated code. In this line of work, a dataset sample is considered correct if it passes a set of unit tests. 

HumanEval \cite{DBLP:journals/corr/abs-2107-03374} is a notable step in this direction. It presents a dataset of hand-written unit tests for programming problems, crafted by programmers, which describes simple program execution, like those in programmers' repositories datasets such as Github. 

At the same time, the granularity of functional correctness in the HumanEval work is limited to the final output of the generated program. In a complex program comprised of multiple steps, we need to provide insights into which parts of the program are correct.  Also, obtaining precise evaluation, especially for complex code with numerous edge cases, requires a comprehensive functional test suite. 
Last, functional correctness metrics are often task-specific. A metric idesigned for one type of task (e.g., generic programming problems)  may not be suitable for another type (e.g., generating code accessing a certain API). 
%
%Without a comprehensive set in place passing all tests doesn't guarantee 100\% accuracy. Yet, Balancing comprehensiveness with scalability in test suites remains a challenge.
%
%Last, there is a risk that models overfit on the provided test cases. If a model is trained or tuned specifically to perform well on known tests, it might not generalize well to real-world, unseen examples.\reut{hmm. do we solve this last bit? if not, remove this sentence?}

%The challenges presented led us to 
To overcome these challenges, in this work we create a  set of functional correctness evaluation tests for programs that originate from novice descriptions. The described scenarios are thus accompanied by comprehensive unit tests, encompassing the various edge cases typical of complex programs, to give us a clear and more granular assessment of the generated code.

%the accuracy in producing control flow structure in the generated code.

\section{Task and Benchmark} \label{section:task}

% \subsection{Task Definition} \label{subsection:task.definition}
%\yg{I think we should at some point mention how we differ from previous works that created APIs "on the fly" to match the kinds of queries as they were annotating the data, e.g. the FB work we looked at in the beginning. We, otoh, have a single, fixed API.} \asaf{In Section 5.2 I specify that the APIs were developed orthogonally to the collected utterances. Do you think we should explicitly have a comparison to these other works?}
%\yg{i think it is a meaningful comparison that plays in our favor, so yes. also, i think 5.2 is a bit too late for that.}\asaf{I emphasized that (now) in the introduction as well}
\paragraph{Input}
The task takes as input (i) a novice-generated natural language utterance, which describes a complex code, and  (ii) API specifications that allow for the functional execution of the relevant code. These API specifications serve as a bridge between the high-level complex NL instructions and the low-level code implementation.%\reut{This whole para was unclear to me} \asaf{my motivation in this para is to give a short context on the grounding of the input. I moved it to the data collection section}

\paragraph{Output}
The output of this task is a program code that satisfies the description given in the NL utterance. The output  code is required to be syntactically and functionally correct and compile successfully
%\footnote{A Python code should interpret correctly.}\reut{didnt understand the footnote purpose}
following the  API specifications. 

\paragraph{Benchmark Creation}
In what follows we discuss the steps towards benchmark creation for the task. First, we collect intent-based program descriptions from non-programmers, serving as a window into the intuitive language constructs and phrasing preferred by this group, capturing a diverse range of plain language descriptions involving different logical control flows (Section \ref{subsection:data.crowd-source}). Second, we create a corresponding wide-coverage, hand-crafted test-suite validating the control flow and edge cases in the described systems, which is used for function-based evaluation (Section \ref{subsection:data.test-suites}). Finally, we create a corresponding code-base solving these utterances, based on a domain-specific API specification which we formulate (Section \ref{section:api-specifications}). %\yg{Note that this is also kind of repetitive with the start of section 4, so need to remove one of them.}

\section{Data Collection and Curation} \label{section:data}

Our benchmark creation begins with collecting natural language utterances from novices, in domains that are understandable to them (Section~\ref{subsection:data.crowd-source}). 
Next, we assemble an evaluation dataset that combines these utterances with corresponding test suites to assess the quality of the generated code (Section~\ref{subsection:data.test-suites}).
In addition, we synthetically generate NL utterance-code pairs to facilitate the training of models for the task (Section~\ref{subsection:synthesized-data}).

\subsection{NL User Requests Collection Interface} 
\label{subsection:data.crowd-source}
The first step of our data collection aims at a scalable collection of natural language instructions from novices that depict complex execution in control flow structures. This collection is achieved via crowd-sourcing. 
We crowdsourced novice descriptions of complex code from crowd-workers\footnote{39 Native English speakers with a high approval rate of above 99\%.} on the Amazon Mechanical Turk platform.
We ground the collected NL utterances in a task-oriented dialogue-systems environment. %\yg{this is a bit weird to have it here, as we already described the API and its domain above. Maybe switch the order of these two sections? and then in the API ones say that the API follows the domain? or some other solution to make this less odd w.r.t the API text}
This realm is a critical component of virtual assistants, which are responsible for understanding the user’s intents (e.g., \textit{set reminder}, \textit{play music}, etc.). It is user-friendly and requires no coding expertise or familiarity with specific programming languages.

We focused our collection efforts on 9 domains and their inherent intents: clock, events, map, messaging, music, reminders, shopping, smart home, and weather. These domains are intuitive to the novice crowd worker, do not require any proficiency, and can be  interleaved to describe the execution of a complex program.
The API covers all these domains, yet it remained hidden from the novice crowd workers, preventing biases that may have influenced their natural language descriptions. This highlights the disconnect between the  novice's phrasing  and the anticipated program.

%\asaf{Start revision (A reviewer asked for more details on the utterances sourcing)}
To stimulate the generation of creative descriptions, the interactive user interface simulates a mobile device. The interface guides the contributors through a series of steps to formulate complex queries, validate the control flow, and submit their utterances. 
To foster creativity, contributors were randomly presented with 
(1) a subset of potential domains, 
(2) a preferred control flow (e.g., \textit{condition} or \textit{sequence}) phrased in an intuitive manner,
%\rt{does a novice know what "a condition" mean?} 
and (3) in certain cases, a restriction on the use of frequently recurring words (e.g., "\textit{if}", "\textit{then}", "\textit{tomorrow}", "\textit{tonight}", "\textit{weather}", "\textit{when}", etc). 

Another tactic employed to foster creativity was providing the task with a single-step non-complex utterance example, randomly selected from the TOP \citep{DBLP:journals/corr/abs-1810-07942} and TOPv2 \citep{DBLP:journals/corr/abs-2010-03546} datasets, and requesting the contributors to rephrase the utterance to express a more complex goal, and educating them what complex utterances might be like.
%\rt{how do they know what "control flow" is? how do we communicte that?}
%\asaf{End revision}

\subsection{Human Crafted Test Suites}
\label{subsection:data.test-suites}
The second stage of our benchmark creation process focuses on the formulation of hand-written test suites. 
For evaluation of the code, we collected manually-crafted Python test suites containing unit tests that are tailored to check the correctness of the code generated for the NL instructions from the first stage. 
%\asaf{Start revision (Reviewer was confused whether the tests were human-generated or had some automation)}
We randomly selected 150 evaluation targets from the 1,200 collected utterances in the first stage (\ref{subsection:data.crowd-source}).
Two undergraduate and graduate computer science students, familiar with Python programming and unit testing, spent about 100 hours constructing the tests to carry out the task reliably.
%\asaf{End revision}

%\asaf{Start revision (Reviewers asked for more data on the unit tests annotators and creators)}
Each annotator coded the functional test with a test scenario snippet, to be seeded with  data used as input to the test, and a set of assertion tests accordingly. 
At runtime, we executed the unit test by combining the test scenario snippet, the generated code, and the assertion tests.
Until the unit test execution, the generated code remains concealed from the programmers coding the unit tests, ensuring an impartial evaluation of the code-generating model.\footnote{To prevent leakage of the problems in the evaluation test suites we archived the plain content of the tests in a gzip file that is not accessible to crawlers or scraping.}%\rt{this para was not very clear to me. what is "seed the data"? what does "we consolidated" mean?}
%\asaf{End revision}

\begin{figure} [htb]
    \begin{lstlisting}[style=python-medium]
# test scenario data seeding
data_model = DataModel(reset=True)
data_sender1 = Contact(text='all
advisors in the committee')
data_model.append(data_sender1)
data_sender2 = Contact(text='all
advisors in the committee')
data_model.append(data_sender2)
data_content = Content(text='confirmation')
data_model.append(data_content)
data_content_neg = Content(text='decline')
data_model.append(data_content)
for sender, content in zip([data_sender1, data_sender2], [data_content, data_content_neg]):
    data_model.append(
        Message(
            sender=sender, 
            content=content
        )
    )
data_event_name = EventName(text='my meeting with them')
data_model.append(data_event_name)
data_model.append(
    CalendarEntity(
        event_name=data_event_name, 
        participants=[data_sender1, data_sender2]
    )
)

# start code block to test
<Generated code embedded here />
# end code block to test

# assertions
test_results = {}

actual = data_model.get_data(CalendarEntity)
expected = [] # expected the program canceled the meeting
entity_assertions(expected, actual, test_results)

assert_test(test_results)
    \end{lstlisting}
    \caption{An example of the unit test code used for evaluating the utterance \textit{"Check that I received confirmation emails from all advisors in the committee or cancel my meeting with them"}. At test time, we embed and execute the generated code within the test framework. This process allows us to evaluate the functional correctness according to the test scenario.}
    \label{fig:enter-label}
\end{figure}

\subsection{Synthesized Training Dataset}
\label{subsection:synthesized-data}

We augment our dataset by following the \citet{berant-liang-2014-semantic} approach to generate synthesized utterances. This approach matches NL sentences with logical predicates. 
First, we sourced a seed lexicon specifying a canonical phrase (e.g., \textit{"check weather"}) for each logical predicate (\texttt{checkWeather}) in the scope of our domains. This lexicon was drawn from a held-out set, a subset of the larger dataset we collected via crowd-sourcing (section \ref{subsection:data.crowd-source}). 

Second, we define a grammar, that along with the seed lexicon and a mock data generator,\footnote{We used \href{https://faker.readthedocs.io/}{Faker}, a Python package that simulates data such as names, addresses, and phone numbers.} can automatically generate a plethora of canonical utterances (\textit{"check the weather in New York City tonight"}) paired with their execution code. 

Having defined the predicates and potential arguments, we expand our grammar with naturally occurring phrases manifesting control-flow rules sourced from the held-out set, facilitating the  generation of control flows with varying degrees of nesting. 

Last, canonical utterances were recursively used to compose complex canonical utterances (e.g. \textit{"check weather tonight and traffic in New York City"}). A generated utterance may not have the elegance of a genuine NL user query, though it still retains the semantics of executed through the code. In contrast to the crowd-sourced user queries, that we allotted for evaluation, the synthesized user queries are merely used as training data to fine-tune large language models. 

Finally, this synthetic process is easily scalable. 
While we focus our synthesized dataset efforts on the specific set of the 9 domains and their inherent intents, adding a domain consists of sourcing a seed lexicon for the new domain, identifying predicates in the domain, and updating the grammar with the new domain grammar rules.

\section{The API Specifications}
\label{section:api-specifications}
%\asaf{Start revision (following reviewers' comments we wanted to provide more clarity on the API)}
%We engineered the API specifications to 
To provide code frameworks that bridge the translation of NL descriptions to their respective code programs, we created an API that generically aligns spans in the user description to code data types and actions.
The generated code must correctly utilize the API endpoints and be executable, allowing us to assess its functionality by executing corresponding tests.
The API specifications followed the nine domain apps defined for sourcing the prompts (see Section \ref{subsection:data.crowd-source}). Nonetheless, the formulation of the API was designed independently from the process of collecting natural language user utterances in these domains.

The API code comprises multiple classes and is designed to provide a comprehensive and modular framework for interacting with the system's functionalities.
% \yg{which classes? maybe list them all in a footnote or in appendix?}
49 interfaces define the different data type entities in our APIs.
%\yg{are these classes or interfaces / dataclasses?} 
\texttt{Location}, \texttt{Contact}, \texttt{DateTime}, are examples of such data types. 
To perform actions with these entities, we expose an additional 11 classes with 34 action methods that are available to be executed. For instance, the \texttt{Messages} class defines methods to send a message and find messages received from specific senders, with desired content, or at a specific time.

Users provide natural language descriptions detailing the actions they want to execute and the specifics of each action. 
The model must interpret descriptions and associate them with the appropriate API classes and methods.

To map from user-specified text spans to domain objects, we assumed the API provides the methods \texttt{resolve\_from\_text} and \texttt{resolve\_many\_from\_text}, which discern relevant spans of word sequences in an utterance and correlate them with specific entities.
For instance, when a user queries for \textit{"the weather on independence day"} the method \texttt{DateTime.resolve\_from\_text} is invoked with the temporal description spans from the request (\textit{"on independence day"}) and returns a \texttt{DateTime} object which is stored in the code. 
%\yg{I eddited a bit, is this correct?}
These variables, representing entities, are then used in other API methods to execute the actions requested in the natural language description (e.g., \texttt{get\_weather\_forecast}). 
It is the API implementor's responsibility to provide suitable implementations for these text-resolution functions. In our evaluation suite, we provide suitable mock implementation (see Section {\ref{subsection:evaluation.simulated-api}}). %\yg{added a bit at the end.}

In certain cases, an entity needs to be inferred from another entity. For such instances, the API offers extra methods such as \texttt{resolve\_from\_entity} or \texttt{resolve\_many\_from\_entity}. 
For instance, the phrase \textit{"After every Astros game"} refers to a set of events on a public calendar. To infer the \texttt{DateTime} entity from each event we iterate on the events and call \texttt{resolve\_from\_entity} on each event (see Figure~\ref{fig:resolve-from-example}).
%\yg{this last bit is not clear to me.}

\begin{figure} [t]
    \begin{lstlisting}[style=python-medium]
events = Calendar.resolve_many_from_text("After every Astros game")
for event in events:
    date_time = DateTime.resolve_from_entity(event)
    content = Content.resolve_from_text("check the traffic")
    Reminders.create_reminder(date_time=date_time, content=content)
    \end{lstlisting}
    \caption{An example demonstrating methods to resolve entities from text spans and other entities in the user utterance \textit{"After every Astros game remind me to check the traffic"}.}
    \label{fig:resolve-from-example}
\end{figure}

%This approach also addresses the cases of co-references where an anaphoric expression is expected to have the same variable as the antecedent. In that case, the generated code is not expected to recover the entity using the \texttt{resolve\_*} methods but reuse an already declared variable.
%\asaf{End revision}

\section{Evaluation} \label{section:evaluation}
%\yg{This section does not defined any metric...}
\subsection{Method}
An essential aspect of any task is its evaluation methodology and thus we introduce an automated execution-based evaluation measure for the proposed \workNameShort task, addressing the challenge we outlined in Section~\ref{section:challenges}.

The evaluation aims to quantify three key factors: (1) The model's efficacy in generating fully executable code, (2) The model's capacity to identify and generate the intended control flows reflected in the natural language utterances, and (3) The success of the model in generating all operations in every execution flow.

Instead of assessing the syntactic correctness of the generated code, we shift the evaluation method towards evaluating functional correctness, wherein a synthesized program is deemed correct if it satisfies a series of unit tests. We build upon this methodology by introducing a collection of human-crafted test suites encompassing a diverse range of test cases, meticulously designed to evaluate the accuracy, robustness, and the generated code logical structure. 
Thus, alongside the NoviCode task, we release our set of 150 hand-crafted evaluation problems.\footnote{The dataset and source code are publicly available at \href{https://github.com/asafam/novicode}{https://github.com/biu-nlp/novicode}}

%\asaf{Start revision (One reviewer whether false positives were identified and whether we planned to switch them with automated tests)}
A human expert manually created each functional test to ensure its accuracy in identifying true positives. %\rt{by humans or automatically? who validated? dont use passive voice..} 
%The expert thoroughly examined edge cases in complex programs, and validated these are covered within the functional tests, particularly focusing on the truth or falsity in conditional control flows. 
To minimize the likelihood of false positives, the expert thoroughly designed multiple test scenarios particularly focusing on the truth or falsity in conditional control flows.%, thereby enhancing the reliability and coverage of the functional tests.%\rt{again, I dont undetstand what you say in this para}
%\asaf{End revision}

\subsection{Metrics}
\label{subsection:evaluation.metrics}
To quantify a model's functional correctness evaluation score, we compute the pass@$k$ \citep{kulal2019spoc} score, employing the method used by \citet{DBLP:journals/corr/abs-2107-03374} and subsequent studies. 
We generate $n \geq k$ samples per task, count the number of correct samples $c \leq n$ which pass the unit tests, and calculate the unbiased estimator: %\reut{can you intuitively explain this term? whats (n-c)? why E? how E is calculated? simple average or something else?}
\begin{equation}
    pass@k := \underset{\text{Problems}}{\mathbb{E}} \left[ 1 - \frac{
        \begin{pmatrix}
        n - c \\ k
        \end{pmatrix}
    }
    {
        \begin{pmatrix}
        n \\ k
        \end{pmatrix}
    }  \right]
\label{equation:pass_at_k}
\end{equation}

%\asaf{Start revision (reviewers asked for more information on the matching of the API resolve\_* methods)}
\subsection{Simulated API Implementation}
\label{subsection:evaluation.simulated-api}
%\yg{why isn't this part of where you describe the evaluation method?}
To facilitate the evaluation by executing the functional correctness tests with the generated code, an implementation of the API was required to be in place. 
In a real-world scenario, a system is required to support the true execution of the API-specified classes and methods. For example, sending emails to selected contacts from the user's address book (e.g. \textit{"mom and dad"}), or retrieving the weather forecast at a specific time and place relative to the user (e.g. \textit{"independence day"} and \textit{"my neighborhood"} respectively).
We provide a mock-up implementation that simulates the proposed actions in the API specifications, allows test input data seeding, and supports evaluating state changes of the underlying data model following the invoked methods.

Another aspect of simulating the API is that it expects an extractive span comparison for instantiating data type entities from the NL description. 
One of the challenges in reliably defining these entities lies in matching the correct span (e.g., \textit{"all advisors in the committee"}) to the expected data dictionary entry (e.g., \textit{"Committee advisors"}). For that, we implemented fuzzy matching over content words in the \texttt{resolve\_*} string comparison methods, ignoring determiners and common prepositions and postpositions. This technique compares the input span with the expected span and produces a BLEU score, accepting a match if it exceeds a specific threshold. We used a BLEU score threshold of 50\% \citep{8813269}.
%\asaf{End revision}

\section{Data Quality Assurance}
%\asaf{Start revision (reviewers asked for more details on the properties of the dataset and its quality)}
\paragraph{NL user utterances collection}
%\yg{the text of this paragraph doesn't match its title}
We implemented a multi-step process to maintain the quality of the collected utterances. 
First, we vetted experienced Amazon Mechanical Turk crowd workers\footnote{Native English speakers with a high approval rate (99\%) and significant experience (over 5,000 completed HITs) on the Amazon Mechanical Turk platform.} contributing to this task to ensure they were non-programmers and had no coding proficiency or educational background in computer science or related fields.
Secondly, we funneled our crowd workers through preliminary qualification tests, to verify they understood the task and its intricacies and could provide a wide range of creative responses, avoiding templated, repeating, and short answers.
Last, we manually reviewed each input and tagged it based on control flow structures it exhibited (i.e., loop, sequence, and condition), as shown in Table \ref{table:data-collection-control-flow}. This process resulted in a collection of 1,200 verified utterances.
%\asaf{End revision}

\begin{table}[t]
\small
\centering
\begin{tabular}{ l c }
\hline
\textbf{Control flow} & \textbf{Frequency} \\
\hline
Sequences   &  57.8\% \\
Conditions  &  31.8\% \\
Loops       &  26.1\% \\
\hline
\end{tabular}
\caption{Frequency of the observed control flows found in the crowd-sourced utterances. Multiple control flows can be seen in a single utterance.}
\label{table:data-collection-control-flow}
\end{table}

\paragraph{Human crafted test suites}
%\yg{shouldn't this come \emph{after} the "Evaluation Metrics" section?}
An expert programmer reviewed each of the 150 selected problems in the evaluation set to ensure the quality of the functional correctness test suites in the benchmark. 
First, by manually providing code for every prompt -- the expert confirmed that every problem had a corresponding code that correctly used the API classes and methods. 
Secondly, to ensure the test is executable and can validate the evaluated code, the expert programmer integrated the manually provided code into the functional correctness tests assigned to the problem and executed it to either pass the test suites upon a correct code or fail otherwise. %\rt{I dont understand what you say from "secondly" to "otherwise"}
This review process resulted in a human performance score of 100\% in the functional correctness evaluation tests. This ensured that all tests were executable upon providing a correct code for the prompt and that the test formulation successfully checked multiple test scenarios.%\rt{which task? writing the code? writing the test suites? and why does it entail 100\%?}
%\asaf{End revision}

\section{Better Code Generation through Intermediary Representation} \label{section:method}
%\yg{this is my attempt to re-write the previous section in  way which is more clear to me.}
Instead of translating natural language utterances directly to code, we propose to map the natural language to an intermediary structure that better encapsulates the control flow elements of the language. 
By explicitly representing the structure of the target complex program through a hierarchical structure, one that expresses the compositional control flows expressed in the code, we hope to improve  the compositional generalization thus 
 improving code generation accuracy  relative to standard end-to-end text-to-code methods \citep{DBLP:journals/corr/abs-2009-06040}.

\begin{figure} [t]
    \begin{lstlisting}[style=python-medium]
[ Module
    [ events = Calendar.resolve_many_from_text('After every Astros game') ]
    [ For
        [ test
            [ iter
                [ events ]
            ]
            [ Name
                [ event ]
            ]
        ]
        [ body
            [ date_time = DateTime.resolve_from_entity(event) ]
            [ content = Content.resolve_from_text('check the traffic') ]
            [ Reminders.create_reminder(date_time=date_time, content=content) ]
        ]
    ]
]
    \end{lstlisting}
    \caption{An example of the \textit{cAST} in a bracket notation.}
    \label{fig:cAST}
\end{figure}

Formally, given a natural language utterance $x$, our goal is to convert $x$ into a surface code $y$. Instead of mapping directly to $y$, we learn to translate $x$ to an intermediary logical form $z$, which can be deterministically converted to $y$. We  use the text-to-code models to perform the transformation $x \rightarrow z$, and then use a deterministic transformation $z \rightarrow y$ to obtain $y$.
Specifically, $z$ is based on a \emph{linearized compact AST} (cAST) which is derived from the abstract syntax tree (AST) of the code, and is linearized into text using a bracketted notation (Figure \ref{fig:cAST}). 
The AST is a tree structure representing the structure of the program. Using the AST allows the model to focus more on the logic and flow of the complex program rather than mere syntax (e.g.,  keywords, symbols, indentation), reducing the complexity of the output space. On top of that, we define a compact  AST (cAST) that only retains the control flow logic structures, such as sequences, conditions, and loops, in a tree form. 
To create compact ASTs, we revert basic operations such as variable assignments or function calls to their fundamental code syntax. These elements are transformed from their AST hierarchical representation to appear as leaves in the tree's code form.

To translate the linearized cAST $z$ back to code $y$, we parse the string into its cAST tree form and expand it back to the AST form using a tool we provide. The ASTs are  translated into actual code using the official Python package for this  (\texttt{ast}).

We note that previous works (\citealp{yin-neubig-2017-syntactic}, \citealp{yin2018tranx}) also utilized ASTs to synthesize code better. Yet, these works model code generation as a series of classification problems of grammar rules required to reconstruct the AST. Our method uses the hierarchical tree form  as a vehicle to directly map texts to hierarchical code structures.

\section{Experiments}
\label{section:experiments}

We set out to evaluate how existing models cope with the \workNameShort task. Moreover, we aim to assess whether our representation-based learning approach (Section~\ref{section:method}) outperforms a simple end-to-end Text-to-Code baseline. 

To do so, we evaluated decoder-only generative large language models (LLMs) using in-context learning scenarios. We also tested pre-trained encoder-decoder LLMs that are fine-tuned with the synthesized trainset. We tested both a simple end-to-end scenario (denoted  \textit{base}) and one where we map to the hierarchical compact AST (denoted  \textit{cAST}), as defined above.
%The various experimental setups are detailed in table \ref{tab:experiment-setup}.

%To assess the impact of our proposed method, we tested conversions from NL utterances (labeled as \textit{text}) to \textit{code}, or to the compacted AST proposed in our structural intermediate code form (referred to as \textit{cAST}).
%These input-output setups we experimented with are shown in Table~\ref{tab:experiment-setup}.

%\begin{table}[t]
 %   \centering   \begin{tabular}{l|cc}         \textbf{Setup} & \textbf{Input} & \textbf{Output}\\       \hline         base &  text & code \\      codeRep & text & cAST \\    \end{tabular}   \caption{The experimental setup of the input options the model took and the output the model was expected to generate}   \label{tab:experiment-setup} \end{table}

\paragraph{Data and Evaluation}
We synthesize a training dataset of 40K samples as discussed in Section~\ref{subsection:synthesized-data}.
Every sample in our dataset is a tuple containing an NL user request, a matching Python code, and the cAST representing the code (Section \ref{subsection:data.crowd-source}). 
We evaluate models on this task using expert-crafted test suites for the functional correction of the generated code (Section \ref{subsection:data.test-suites}).

\paragraph{Metrics}
To reliably estimate the functional correctness of a model, we generated 200 samples for each prompt ($n = 200$) and calculated the mean $pass@1$ and $pass@10$ scores.

\paragraph{In-context Learning.}
\label{subsection:experiments.incontext}
We conducted tests on OpenAI's GPT-3.5-Turbo\footnote{\textit{gpt-3.5-turbo-1106}}, and GPT-4-Turbo\footnote{gpt-4-0125-preview} models using  in-context learning (prompting)  \citep{brown2020language}.
%\asaf{Start revision (Reviewers asked for results on open source LLMs and also some more std dev stats on the results)}
We also experimented with open source LLMs including Meta's CodeLlama,\footnote{\textit{CodeLlama-7b-Instruct-hf}} Mistra,l\footnote{\textit{Mistral-7B-Instruct-v0.2}} and DeepSeek Coder.\footnote{\textit{deepseek-coder-33b-instruct}}
%\asaf{End revision}
Our experimental setup used in-context prompts that included the full API specifications and multiple examples of few-shot prompts randomly selected from the synthetically generated  dataset.
The examples count was capped at 11, aligning with the smallest context window of our models, which equals 16,385 tokens.
This type of in-context prompting was feasible only with models that have a higher token limit for context windows, as %\asaf{Start revision (relates to previous LLM experiment request)}
the API code itself contained 15,397 tokens (including the Python Docstrings).
%\asaf{End revision}

\paragraph{Fine-tuned Models.}
We further assess the NoviCode task by  fine-tuning models that were previously shown to  successfully perform program-synthesis tasks. These models include T5 (with 220M parameters) \citep{t5}, CodeT5 (220M) \citep{wang2021codet5}, and CodeT5+ (220M) \citep{wang2023codet5+}. Each model was fine-tuned using input-output tuples tailored to the specific experimental setups, utilizing data from the synthesized training dataset. The dataset of synthesized samples was split into train (80\%), validation (10\%), and test (10\%) sets for the fine-tuning phase. All models were configured with a maximum input and output length of 512 tokens each. A learning rate of 5e-5 was employed, alongside a constant warm-up with step inverse decay, and the warm-up steps were capped at 1,000. We utilized the AdamW optimizer \citep{loshchilov2019decoupled}, and the experiments were conducted with a batch size of 8 and were set to run for a maximum of 20 epochs. However, the execution was halted if no improvement was observed for three consecutive epochs. An A100 GPU machine was employed for these experiments.

\section{Results and  Analysis} 
\subsection{Results}
\label{subsection:results}

Our experiments revealed that in most model architectures we tested, having an output of a compacted AST (cAST) structural form showed increased performance compared to the text-to-code method. 
The best results were obtained with the GPT-4-Turbo model, to which we supplied an in-context prompt containing the NL descriptions and expecting a \textit{cAST} to be converted to code.

%\asaf{Start revision (following reviewr's comment on doing some more analysis on cAST success)}
Analyzing the results, we observed that the \textit{cAST} form better generated code that exhibited conditions and loops in control flows, forms which are more explicit in the ASTs. 
With sequences of operations, the two setups showed similar success. In cases where multiple control flows were present in the same utterance, the \textit{cAST} output form also excelled compared to the text-to-code setup.
%\asaf{End revision}

\paragraph{In-Context Learning} 
%We assessed this task on different OpenAI GPT LLMs. 
%The setup we first used included in-context prompts and expected the model to generate a Python code output (see Section~\ref{subsection:experiments.incontext}).

%The results of our in-context experiments on various OpenAI GPT models are shown in Table \ref{table:experiments.incontext}.
%In-context prompts that contained the complete API specification (supported only in newer models with larger context windows) showed a notable increase in functional correctness scores across the evaluation set relative to providing examples only. 

%When evaluating the standard text-to-code approach against a setup of our proposed structural representation code form we used the most effective OpenAI model identified in earlier experiments (GPT-4-Turbo with a 128K token context window). The results are shown in Table \ref{table:experiments.incontext}. Our proposed code representation method showed the best result. This performance in an in-context learning format, especially considering that the GPT-4-Turbo was trained on text-to-code datasets, shows the promise of using our method.

We assessed this task on different commercial and open-source LLMs. 
Our assessment evaluated the standard text-to-code approach against our proposed structural representation code setup.
The results are shown in Table \ref{table:llm-experiments-cv}.
Our proposed approach using the code representation method showed the best result. 
This performance in an in-context learning format, shows the promise of using this intermediate form to represent  code in general. 

\begin{table*}
\footnotesize
\centering
\begin{tabular}{ l l l l c c}
\hline
\textbf{Model Name} & \textbf{Setup} & \textbf{pass@\textit{1}} & \textbf{pass@\textit{10}}\\
\hline
\multirow{2}{*}{GPT-4-Turbo} & base & $33.8 \pm 0.1$ & $51.6 \pm 0.4$ \\
                             & cAST & $\mathbf{39.0 \pm 0.3}$ & $\mathbf{55.6 \pm 0.8}$ \\
\hline
\multirow{2}{*}{GPT-3.5-Turbo} & base & $10.7 \pm 0.0$ & $29.0 \pm 0.0$\\
                               & cAST & $10.3 \pm 0.1$ & $31.3 \pm 0.2$ \\
\hline
\multirow{2}{*}{CodeLlama-7B} & base & $1.5 \pm 0.1$ & $10.7 \pm 2.1$ \\
                              & cAST & $3.3 \pm 0.0$ & $17.2 \pm 1.5$ \\
\hline
\multirow{2}{*}{Mistral-7B} & base & $0.3 \pm 0.0$ & $2.1 \pm 0.3$ \\
                            & cAST & $1.2 \pm 0.3$ & $6.6 \pm 1.2$ \\
\hline
\multirow{2}{*}{DeepSeek-Coder-33B} & base & $8.8 \pm 0.3$ & $27.8 \pm 0.2$ \\
                               & cAST & $7.0 \pm 0.4$ & $26.2 \pm 0.3$ \\
\hline
\end{tabular}
\caption{Comparing LLMs with different input-output setups. We find that representing code in a structural form outperforms the basic text-to-code approach. Results are indicated by mean and std dev.}
\label{table:llm-experiments-cv}
\end{table*}

\paragraph{Fine-tuned Models}
We turned to assess this task on three fine-tuned baseline models. 
The results are presented in Table~\ref{table:fine-tune-modesl-results}.
%These models were fine-tuned with our text-to-hierarchical approach.
 In the pass@$k$ scores, the hierarchical code representation method (using cAST) outperformed all text-to-code strategies in all model architectures we tested. 

\begin{table}[ht]
\footnotesize
\centering
\begin{tabular}{ l l c c }
\hline
\textbf{Model Name} & \textbf{Setup} & \textbf{pass@\textit{1}} & \textbf{pass@\textit{10}} \\
\hline
\multirow{2}{*}{T5} & base    & 0.0 $\pm 0.0$ & $0.0 \pm 0.0$ \\
                    & cAST    & $9.7 \pm 1.1$ & $\mathbf{19.6 \pm 2.3}$ \\
\hline
\multirow{2}{*}{CodeT5} & base &  $8.1 \pm 0.7$ & $14.8 \pm 1.6$ \\
                        & cAST &  $9.1 \pm 0.5$ & $18.8 \pm 1.1$ \\
\hline
\multirow{2}{*}{CodeT5+} &  base  &  $11.3 \pm 1.4$ & $19.4 \pm 2.8$ \\
                         & cAST   &  $\mathbf{11.7 \pm 0.8}$ & $18.8 \pm 2.1$ \\
\hline
\end{tabular}
\caption{Comparison of fine-tuned models' performance. Our proposed hierarchical code representation form achieved the highest performance scores across all models. Results are indicated by mean and std dev.}
\label{table:fine-tune-modesl-results}
\end{table}

\begin{figure}[t]
    \centering
    \includegraphics[width=0.5\textwidth]{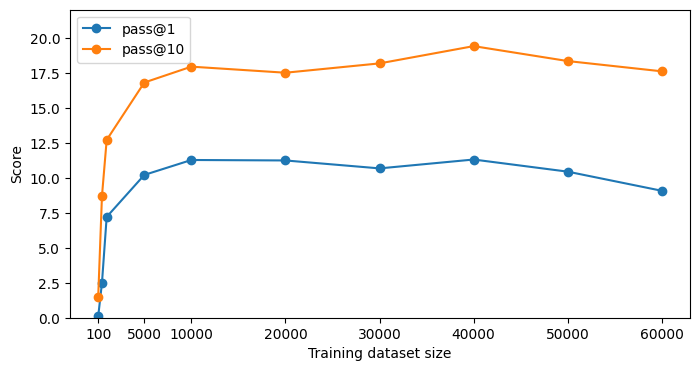}
    \caption{Comparison of fine-tuned models' performance for varying dataset sizes. The success of our proposed approach for this task is limited by the reliance on synthesized examples for fine-tuning models.}
    \label{fig:synthesize-dataset-size}
\end{figure}

\paragraph{Training Size Learning Curve} 
To understand the minimal train-set size required for fine-tune a model using our approach, we evaluated the best-performing CodeT5+ model with the proposed structural code representation approach, by incrementally increasing the train set used for fine-tuning.
We observe that models trained on smaller datasets (fewer than 5,000 samples) yielded notably lower scores (Figure~\ref{fig:synthesize-dataset-size}). 
As the train set expanded, the performance improved consistently and then reached a plateau. %, achieving a maximum of 13.5 pass@$1$ and 20.9 pass@$10$ scores. 
%This suggests that simply increasing the synthetic training-set volume may not be a promising factor for further improvement. This could be due to models overfitting over synthesized samples in the training set, which are often derived from templates and predetermined grammar rules. 
%Furthermore, 
This implies manually sourcing a sufficiently large dataset would be difficult because of the substantial volume needed. %\reut{hypothesize why. overfit synthetic patterns? away from real distribution?}

\subsection{Error Analysis}
We conducted an in-depth error analysis of the results of the best-performing model (GPT-4-Turbo with a 128K token context window using the \textit{cAST} in-context setup) to identify key areas for improvement. We based our analysis on a randomly selected subset of 150 outputs generated by the model.

Observing the model's efficacy in generating code programs with accurate control flows (Table~\ref{table:control-flow}),
our analysis revealed that control flows that are more explicitly expressed in the NL description, such as \textit{sequences} and \textit{conditions} (e.g., conditions that were described using the function word \textit{if}) were more accurately implemented in code.  Loops, in contrast ,often require implicit deduction from nuances like quantifiers (e.g., \textit{every day}), noun phrases with conjunctions (e.g., \textit{mom and dad}), or specific semantic terms (e.g., \textit{my book club group}), and are harder to infer.
%Sequences of executions in code, being the most basic control flow structures, are generally captured more successfully - in 68\% of the cases. 
%Our analysis also revealed that conditions are more accurately generated compared to loops. This aligns with expectations, as conditions in NL descriptions are usually explicitly expressed using function words (e.g., \textit{if}), whereas loops often require implicit deduction from nuances like quantifiers (e.g., \textit{every Starbucks}), noun phrases with conjunctions (e.g., \textit{Monday and Tuesday}), or specific semantic terms (e.g., \textit{my book club group}).

\begin{table}[t]
\small
\centering
\begin{tabular}{ l c }
\hline
\textbf{Control Flow} & \textbf{Success \%} \\
\hline
Sequences   &  68\% \\
Conditions  &  46\% \\
Loops       &  16\% \\
\hline
\end{tabular}
\caption{Model success in recovering complex descriptions to program code with control flows. The more explicit control flow structures appear in the NL description -- the better the model can embed it in code. }
\label{table:control-flow}
\end{table}

The model displays distinct patterns of errors that provide insight into its limitations and potential areas for improvement. We broadly classified errors into three categories: 
(i) Syntactic errors: malformed cAST 
(ii) Logical errors: Compiled code with runtime errors/exceptions
and (iii) Semantic errors: Successful run with a wrong outcome. We hereby detail them in turn.
%See appendix \ref{appendix:error-analysis} for examples.\reut{no appendix..}
 
\paragraph{Syntactic errors.}
These were rare, seen in only 7.2\% of the cases, but critical, as the model generated code  with incorrect syntax, such as malformed AST node labels or mismatched brackets. In that case, we were not able to reconstruct the final Python code from the code intermediate representation as cAST. 
A related case is where the model generated an (too-) lengthy output and reached its maximum token limits, neglecting the correct closure of brackets in the linearized tree.

\paragraph{Logical errors.}
In 26\% of the cases, we noted runtime errors where, although we successfully transformed the intermediate representation into a program code, it led to exceptions upon execution. Examples of these errors include referencing undefined variables, incorrectly calling functions with unexpected arguments, or illegal operations on data types, like attempting to iterate over objects that are not iterable.

\paragraph{Semantic errors.}
These were the most common (53\%), where the model succeeded in executing the generated code but failed to implement correctly the NL description, leading to a completely different output than expected. Semantic errors were detected following assertion failures as part of the functional correctness evaluation (Sec~\ref{subsection:data.test-suites}).
This error type is manifested in outputs containing irrelevant or incorrect data not present in the NL input, poor recall in identifying necessary arguments due to data type errors or omissions, and problems with variable reusability, especially in cases where  antecedents and anaphors are distanced in the NL descriptions.

\section{Related Work}

%\paragraph{Datasets for Code Related Tasks.}
Previous text-to-code datasets were delivered to facilitate the training and testing of the code generation capacity of contemporary models. Notable datasets include the CoNaLa Dataset \citep{DBLP:journals/corr/abs-1805-08949}, which contains programming questions from Stack Overflow along with code solutions, the CONCODE dataset \citep{DBLP:journals/corr/abs-1808-09588}, the CodeSearchNet Corpus \citep{DBLP:journals/corr/abs-1909-09436}, and the CodeXGLUE dataset \citep{lu2021codexglue}, constructed using code-comment pairs from GitHub across numerous domains in various programming languages. The majority of these include very technical jargon and relatively short statements, in contrast with our novice users language and lengthier intent expressions  %(We discuss the shortcomings of these datasets in
(Section~\ref{section:introduction}).
%

%\paragraph{Task-Oriented Datasets.}
Identifying instructions in Task-Oriented datasets can also be seen as equivalent to interpreting description as executions. Notable Task-oriented datasets are TOP \citep{DBLP:journals/corr/abs-1810-07942} and TOPv2 \citep{DBLP:journals/corr/abs-2010-03546} datasets. However, these datasets fail to provide high-level abstractions that elicit complex code with sequences and control flow structures (e.g., loops and conditions). 

%\paragraph{Automatic Evaluation Metrics.}
On the front of evaluating text-to-code generation, automatic metrics for code generation evaluation initially adopted techniques similar to those used in machine translation. 
BLEU \citep{papineni-etal-2002-bleu} has been  extensively used for evaluating code generation. However, its limitations have been increasingly recognized, including the disregard for semantic correctness and functionality. 
CodeBLEU \citep{DBLP:journals/corr/abs-2009-10297} enhances BLEU by incorporating crucial code-related features such as syntactic and semantic similarity, data flow, and variable misuse. Despite its improvements, CodeBLEU still relies on a reference implementation, which may hinder its efficacy in cases where multiple correct solutions exist. 

The HumanEval dataset \citep{DBLP:journals/corr/abs-2107-03374} presented a recent and significant effort in this direction, where functional correctness tests evaluate the generated code. The evaluation tests in HumanEval present basic programming problems.
%\asaf{Start revision}
\citealp{austin2021program} presents a similar approach for evaluating correctness. 
Additional code generation benchmarks extended this approach using more challenging programming problems, which require an understanding of algorithms (\citealp{Li_2022_AlphaCode}, \citealp{hendrycks2021measuring}), or usage of external and advanced Python packages (\citealp{lai2022ds1000}, \citealp{wang2023executionbased}). 

Our benchmark presents code-generation tasks that concentrate on everyday tasks for laypeople, such as scheduling meetings in a calendar or checking weather forecasts. Furthermore, to support and extend the complexity, our benchmark introduces a private and unseen API, in contrast to using standard Python packages.

Other benchmarks challenged models by providing class-level Python code-generation tasks \citep{du2023classeval}. 
The challenge introduced in this paper is orthogonal to this challenge and does not generate class-level code.%\rt{huh? we do not generate functions?} 

Functional tests on the generated code are prone to false positives or true negatives upon low coverage of test edge cases. Some works \citep{liu2023code} mitigate this using an automatic test input generation engine. In contrast with this approach, our evaluation method validates the data model affected by the generated code and not the generated code itself.

Few code generation benchmarks have also concentrated on the population composing the NL prompts in the benchmarks for code LLMs, specifically targeting non-expert beginner programmers (\citealp{babe2023studenteval}).
In contrast, the prompts in our benchmark were crafted by novice, non-programmers lacking any programming proficiency.

Last, other corpora capture how non-programmers express if-then clauses \citep{quirk-etal-2015-ifttt}. Yet, the tasks in this benchmark are expressed in a highly structured and noisy language and exhibit only a single control flow paradigm -- conditionals.

\section{Conclusion}

In this paper we introduce \workNameShort, a novel NL programming task of generating executable complex programs with control flow structures from novice NL user requests and  API specifications that the program should comply with. As an integral part of our task, we propose an evaluation framework to assess model efficacy on this task, based on functional execution and denotation rather than on the code form. Building upon this task, we propose a novel representation that explicitly reflects the hierarchical structure of code, which outperforms all baseline models, open source (e.g., CodeT5+) or closed source (e.g., GPT-4-Turbo) models, in the evaluation tests. 
Finally, our analysis suggests that while generating working code based on language is a feasible task, ours is still a challenging one. Yet NoviCode constitutes a promising approach towards true {\em natural language programming} --- where humans program in their native tongues ---encouraging future research and development of this domain.

\section*{Limitations}

\paragraph{Evaluation Test Size.}
Creating unit test suites for evaluating code generation models on this task is both time-intensive and requires Python testing skills and familiarity with specific APIs. This process significantly contributed to the development time of our evaluation dataset. On average, it took an experienced programmer about 9 minutes to create each unit test. Due to limited resources, we could only prepare tests for 150 out of the 1200 collected user utterances. Expanding the dataset for a more extensive evaluation is a key goal for future work.

\section*{Acknowledgements}

We gratefully acknowledge the contribution of Tamar Gur for her invaluable assistance in this work. 
Additionally, we extend our gratitude to Royi Rassin, Royi Lachmy, Avshalom Manevich, and Shira Kritchman for their helpful comments and discussions.
We also thank the anonymous reviewers and the action editor for their valuable suggestions.

This project received funding from the European Research Council (ERC) under the European Union’s Horizon 2020 research and innovation program, grant agreement No. 677352 (NLPRO) and grant agreement No. 802774 (iEXTRACT).

\bibliography{tacl2021, custom}
\bibliographystyle{acl_natbib}

\appendix

\section{Appendix: Code Intermediate Representation Scheme}
\label{appendix:code-rep}

The intermediate form for representing the code used Abstract Syntax Tree (AST) formulation as its baseline.

Let $T$ be a tree, and $N$ be the set of all nodes in $T$. 

Let $n1 \in N$ be a node in $T$. 

We express that the label of a node \( n1 \) as in the set of allowed labels \( L \) as follows:

\[ \text{lab}(n1) \in L \]

Let \( T_{n_1} \) denote a subtree with \( n_1 \) as its root. Given that the terminals of \( T_{n_1} \) are \( t_1, t_2, \ldots, t_n \) in order, the label of \( T_{n_1} \) can be expressed as the concatenation of labels of the terminals:

\[ \text{lab}(T_{n_1}) = \text{lab}(t_1) || \text{lab}(t_2) || \ldots || \text{lab}(t_n) \]

We say that a subtree \( T_{n_1} \) starts with a string \( s \) if and only if:

\[ s \sqsupset \text{lab}(T_{n_1}) \]

\subsubsection*{Compactization rule}

Let $L = \{'Assign', 'AugAssign', 'AnnAssign', \\'Call', 'Expr'\}$. And let the label of a node \( n1 \) in the set of allowed labels \( L \) as follows:

\[ \text{lab}(n1) \in L \]

We say that a subtree \( T_{n_1} \) should be replaced with a node with the label of the unparsed value of that node.

\begin{center}
\begin{tikzpicture}[level distance=0.75cm, font=\footnotesize,
    highlighted/.style={rectangle,red,fill=blue!15},
    special/.style={rectangle,blue,fill=red!15}]
    \node {n1}
        child{node {*}}
    ;
\end{tikzpicture}
\end{center}

Translates as follows:\\

\begin{center}
\begin{tikzpicture}[level distance=0.75cm, font=\footnotesize,
    highlighted/.style={rectangle,red,fill=blue!15},
    special/.style={rectangle,blue,fill=red!15}]
    \node {$unparse(n1)$}
    ;
\end{tikzpicture}
\end{center}

\section{NL User Utterance Elicitation Interface}

\begin{figure}[ht]
    \centering
    \includegraphics[width=0.5\textwidth]{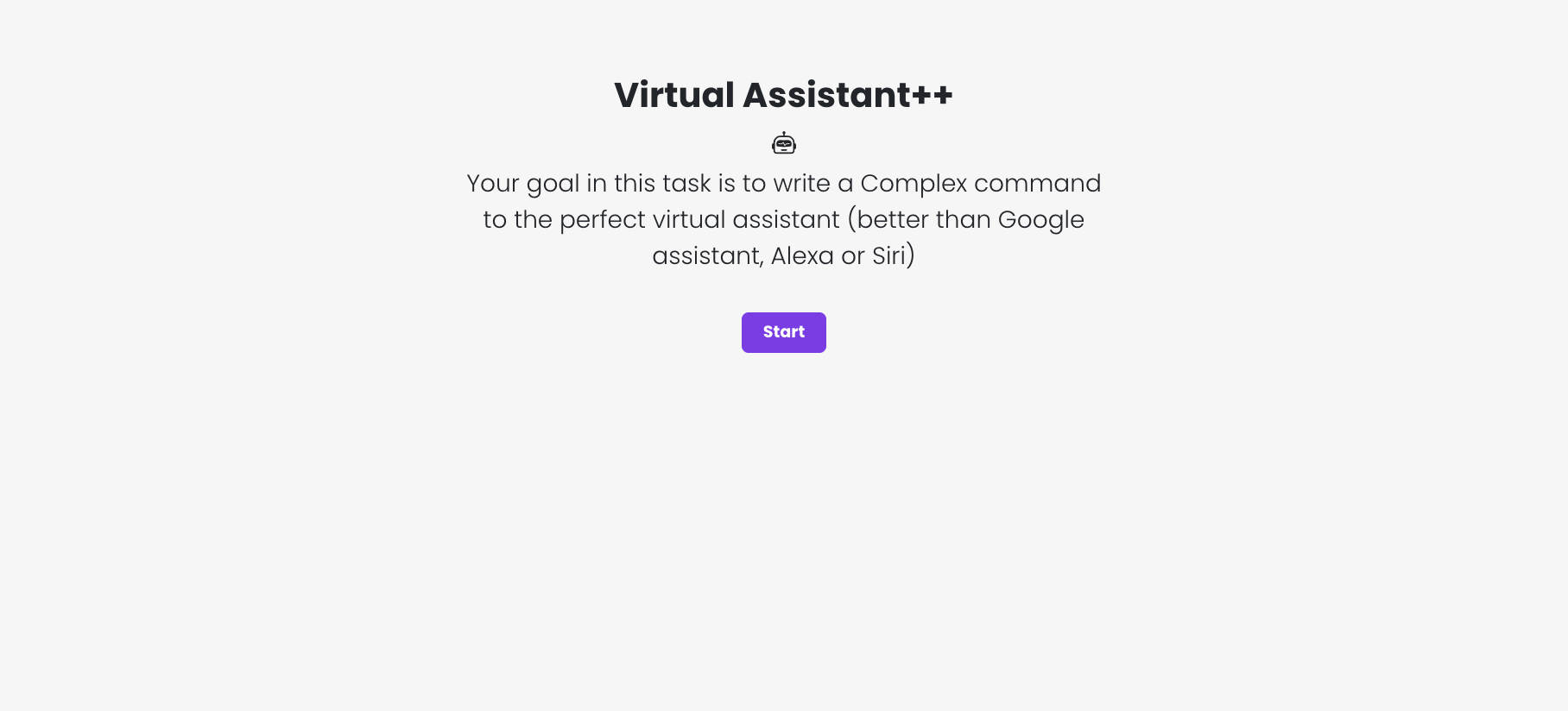}
    \caption{Task introduction screen}
\end{figure}

\begin{figure}[ht]
    \centering
    \includegraphics[width=0.5\textwidth]{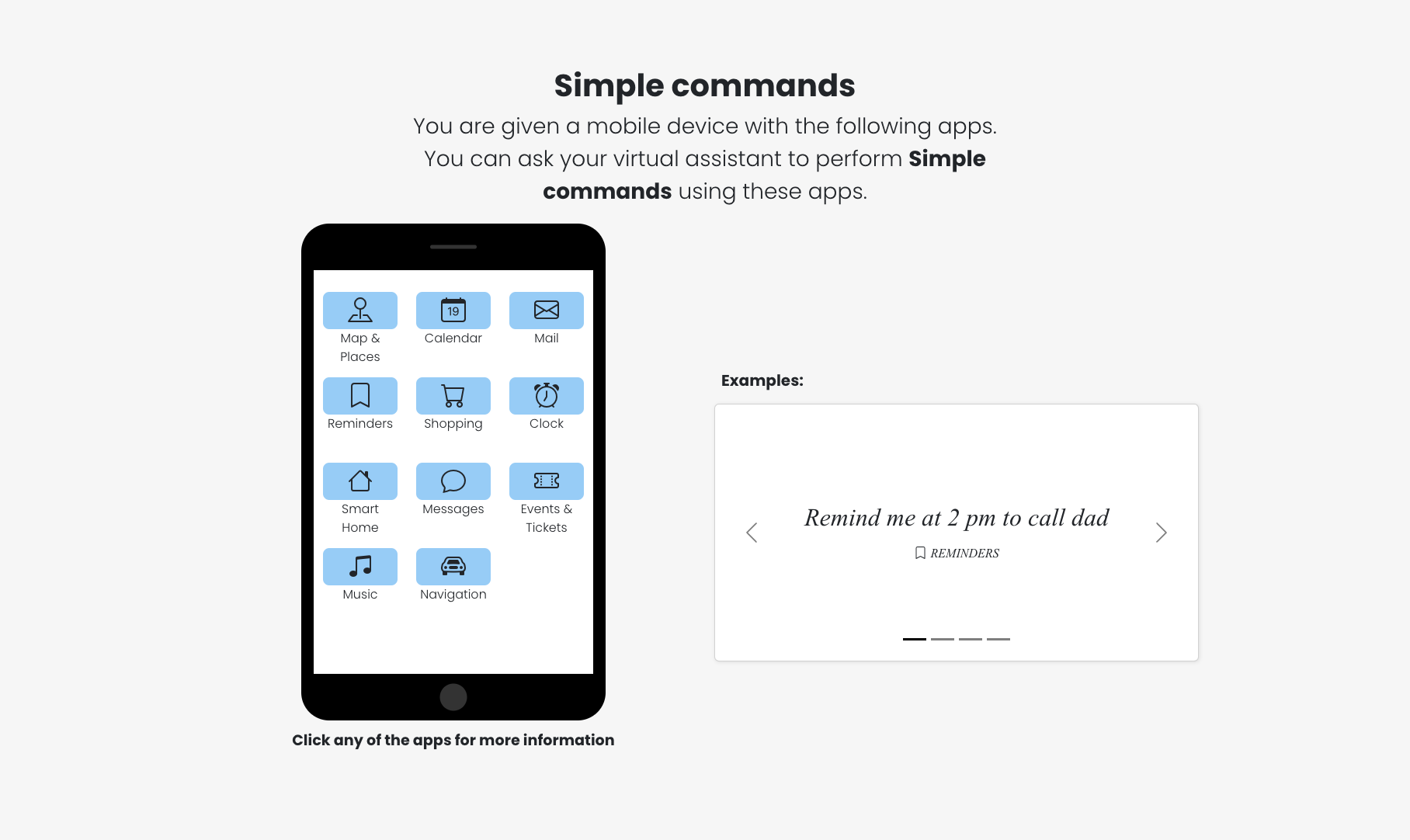}
    \caption{Educating crowd workers on simple instructions}
\end{figure}

\begin{figure}[ht]
    \centering
    \includegraphics[width=0.5\textwidth]{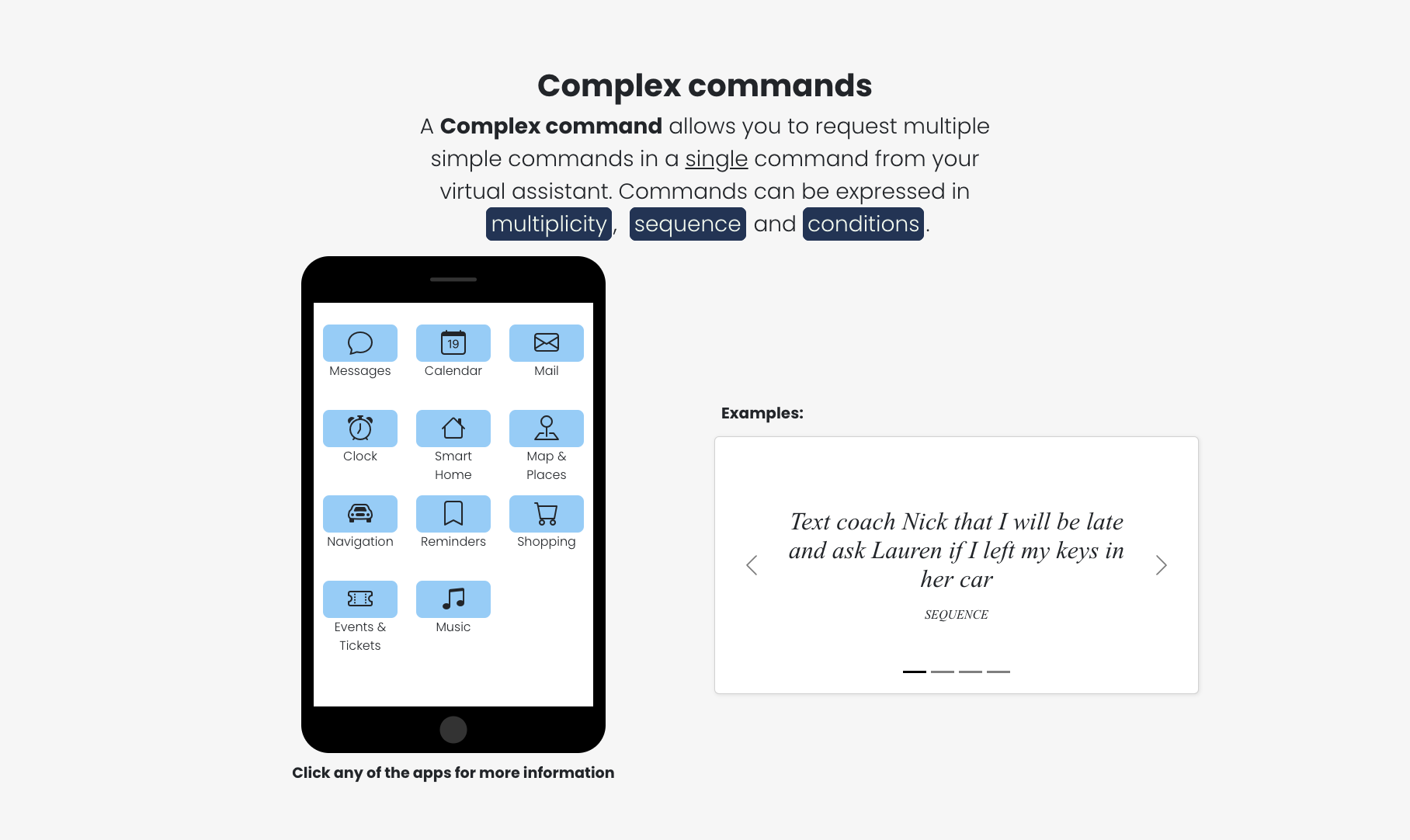}
    \caption{Educating crowd workers on complex instructions}
\end{figure}

\begin{figure}[ht]
    \centering
    \includegraphics[width=0.5\textwidth]{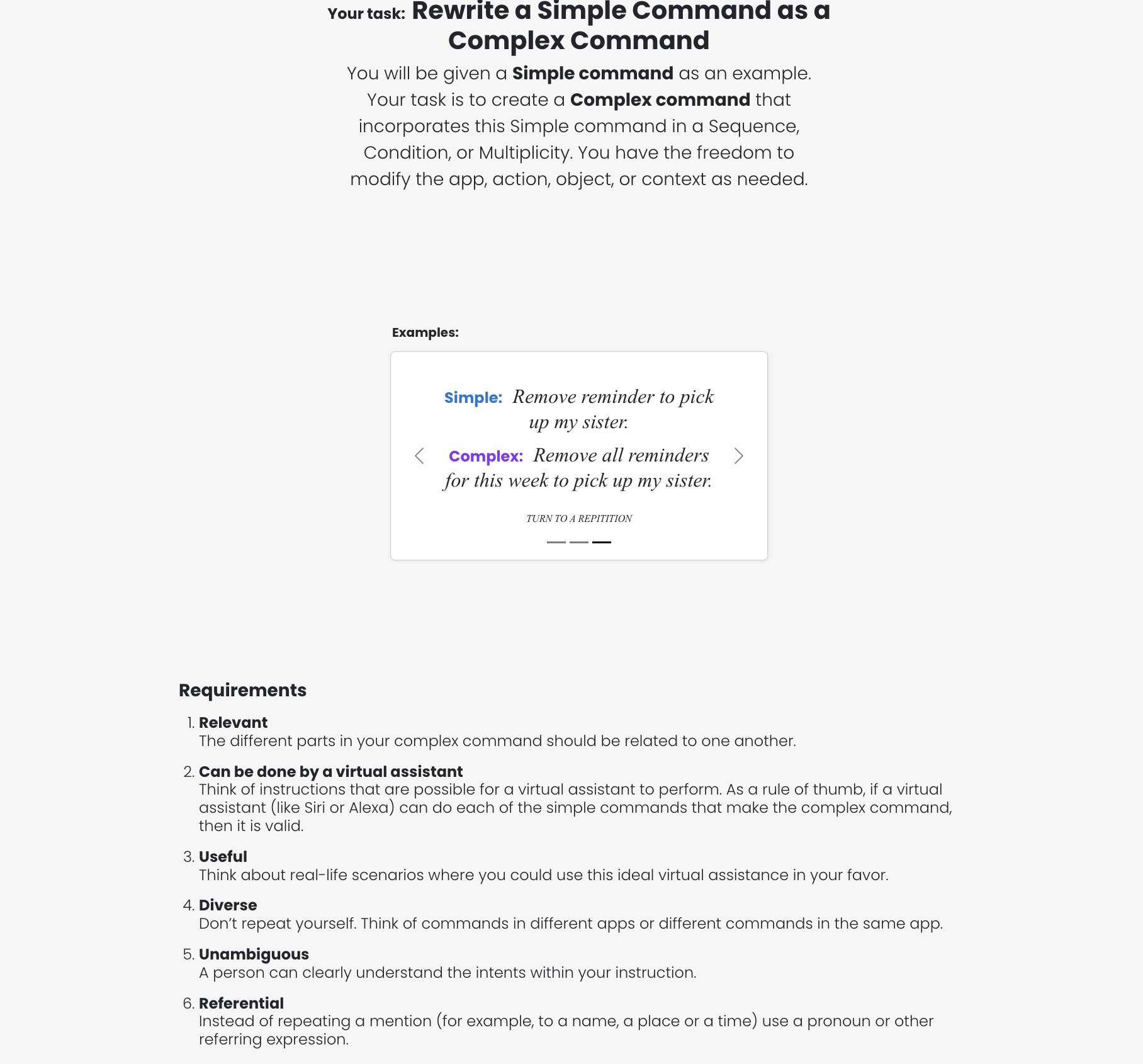}
    \caption{Goals and instructions}
\end{figure}

\begin{figure}[ht]
    \centering
    \includegraphics[width=0.5\textwidth]{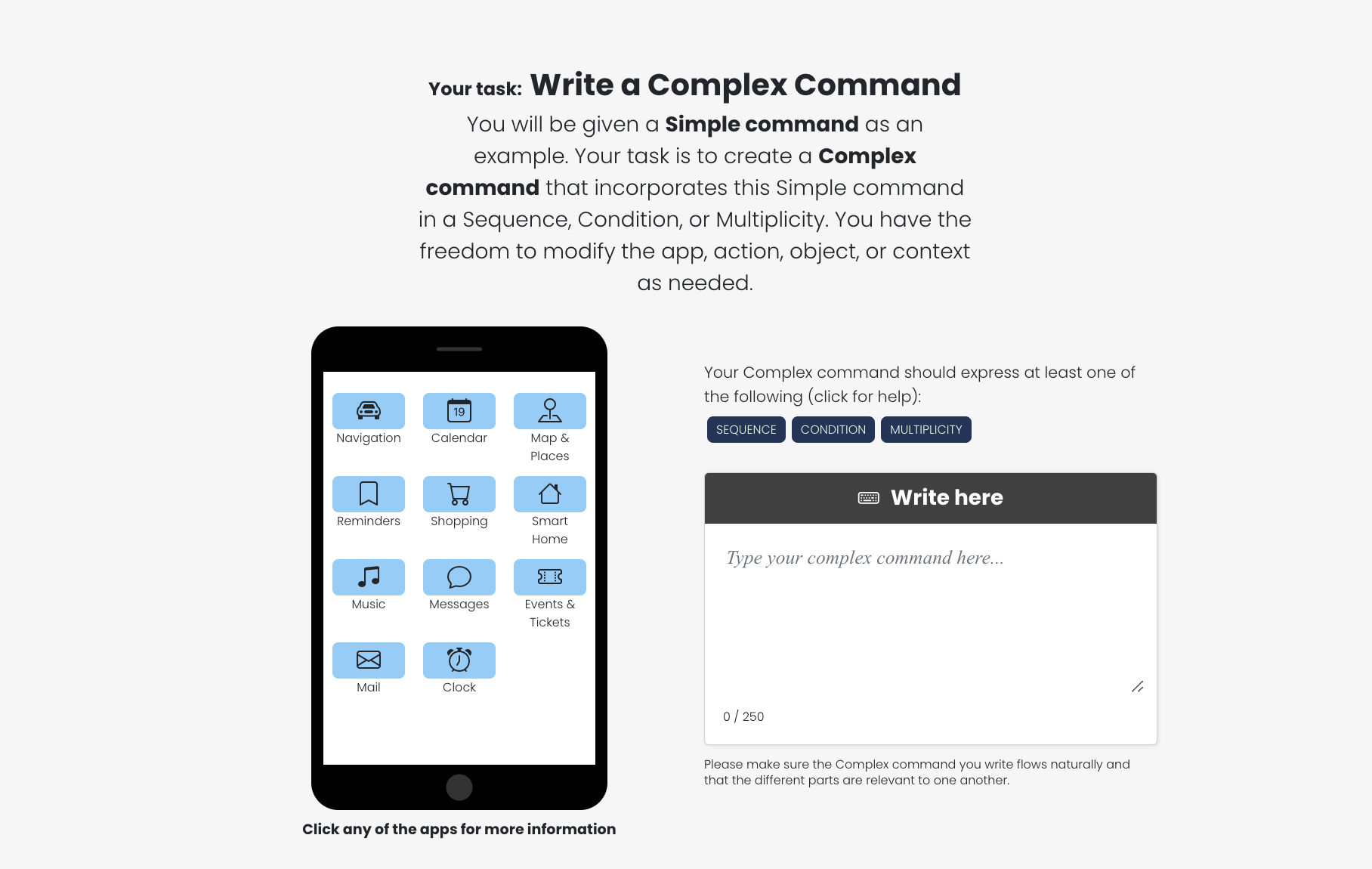}
    \caption{Task input form}
\end{figure}

\begin{figure}[ht]
    \centering
    \includegraphics[width=0.5\textwidth]{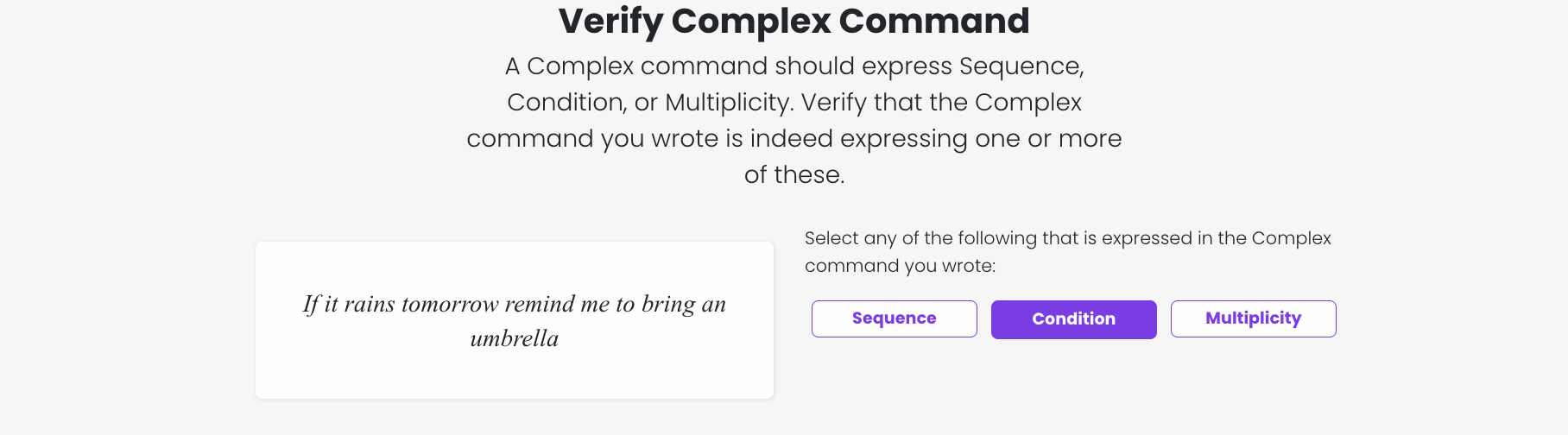}
    \caption{Response verification}
\end{figure}

\begin{figure}[ht]
    \centering
    \includegraphics[width=0.5\textwidth]{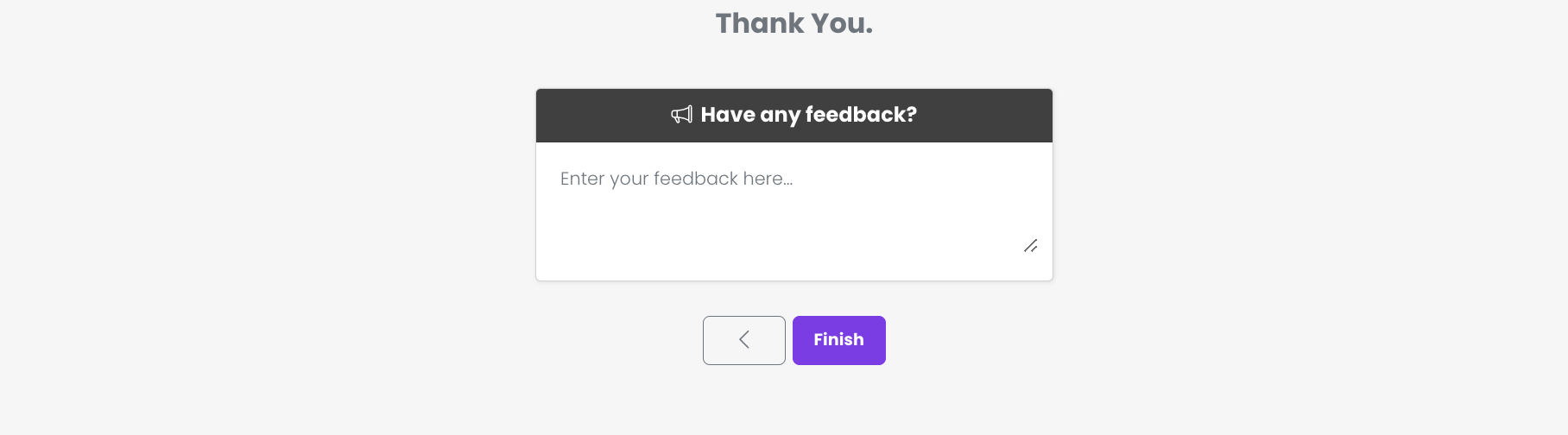}
    \caption{Feedback screen}
\end{figure}

\end{document}